\documentclass{article}
\usepackage{spconf,amsmath,graphicx}

\usepackage{times}
\usepackage{epsfig}
\usepackage{graphicx}
\usepackage{amsmath}
\usepackage{amssymb}
\usepackage{multirow}

\usepackage{microtype}
\usepackage{graphicx}
\usepackage{subfigure}
\usepackage{booktabs} 
\usepackage{amsthm,amsmath,amssymb}
\usepackage{mathrsfs}
\usepackage{bbm}
\usepackage{stfloats}
\usepackage{graphicx} 
\usepackage{float} 
\usepackage{subfigure} 
\usepackage{hyperref}
\usepackage{stfloats}
\usepackage{color}
\usepackage{algorithm}
\usepackage{algorithmicx}
\usepackage{algpseudocode}
\usepackage{textcomp}
\usepackage{amsmath}
\usepackage{hyperref}

\title{Deep Manifold Graph Auto-Encoder for Attributed Graph Embedding}
\name{Bozhen Hu$^{1, 2}$, Zelin Zang$^{2}$, Jun Xia$^{2}$, Lirong Wu$^{2}$, Cheng Tan$^{2}$, Stan Z. Li$^{2 \star}$\thanks{`$\star$' denotes the corresponding author.}}
\address{$^{1}$ Zhejiang University, Hangzhou, 310058, China\\
$^{2}$AI Division, School of Engineering, Westlake University, Hangzhou, 310030, China}

\begin{document}
%
\maketitle
\begin{abstract}
Representing graph data in a low-dimensional space for subsequent tasks is the purpose of attributed graph embedding. Most existing neural network approaches learn latent representations by minimizing reconstruction errors. Rare work considers the data distribution and the topological structure of latent codes simultaneously, which often results in inferior embeddings in real-world graph data. This paper proposes a novel Deep Manifold (Variational) Graph Auto-Encoder (DMVGAE/DMGAE) method for attributed graph data to improve the stability and quality of learned representations to tackle the crowding problem. The node-to-node geodesic similarity is preserved between the original and latent space under a pre-defined distribution. The proposed method surpasses state-of-the-art baseline algorithms by a significant margin on different downstream tasks across popular datasets, which validates our solutions. We promise to release the code after acceptance.
\end{abstract}
\section{Introduction}
\label{sec:intro}

Attributed graphs are graphs with node attributes/features that emerge in various real-world applications. Such examples include social networks~\cite{hastings2006community}, citation networks~\cite{kipf2016semi}, protein-protein interaction networks~\cite{neyshabur2013netal}, etc. From a data representation perspective, graph embedding is to encode the high-dimensional, non-Euclidean information about the graph structure and attributes associated with the nodes and edges into a low-dimensional embedding space. The learned embeddings can then be used for various data mining tasks, including clustering~\cite{wang2017mgae,DCRN}, recommendation~\cite{wang2017mgae} and prediction~\cite{kipf2016variational,JunXia2023SimGRACEAS,xia2022progcl}.

Early graph embedding approaches are based on Laplacian eigenmaps~\cite{newman2006finding}, matrix factorization~\cite{li2018community}, and random walks~\cite{perozzi2014deepwalk}. Recently, there has been a surge of approaches focusing on deep learning methods on graphs. Specifically, Graph Convolutional Network (GCN) based methods such as graph auto-encoder (GAE) and variational graph auto-encoder (VGAE)~\cite{kipf2016variational}, AGC~\cite{zhang2019attributed}, DAEGC~\cite{wang2019attributed} and ARGA~\cite{pan2019learning} obtain embeddings by enforcing the reconstruction constraint, have made significant progress in many graph learning tasks~\cite{zhou2020graph}. Moreover, GraphMAE~\cite{ZhenyuHou2022GraphMAESM} focuses on feature reconstruction with a masking strategy to increase its robustness. The task of most existing deep learning methods requires preserving information beneficial to reconstruction and fails to optimize embeddings directly to maintain the topological structure information yet gets inferior representation in many cases. Worse, when the high-dimensional data is mapped into a low-dimensional latent space, a crowding problem~\cite{van2008visualizing} may occur. In detail, the embeddings generated from such methods are insufficiently constrained, resulting in the appearance of curvature folds that do not exist in the original data space and eventually distorting the embeddings, which means nodes of the same class are prone to crowd together. 

Manifold learning unveils low-dimensional structures from the input data based on the manifold assumption, i.e., data lie on a low-dimensional manifold immersed in the high-dimensional ambient space. Equipped with deep neural networks, deep manifold learning (DML) is transferable to graph data, e.g., MGAE~\cite{ChunWang2017MGAEMG} advances the auto-encoder to the graph domain to embed node features and adjacency matrix. Inspired by MGAE and DLME~\cite{ZelinZang2022DLMEDL}, DML can be applied in learning graph embeddings while keeping distances between nodes. Different from them, we relieve and even tackle the crowding problem by preserving the topological structure for latent embeddings of the graph data under a distribution efficiently. Therefore, we propose the Deep Manifold (Variational) Graph Auto-Encoder (DMVGAE/DMGAE) method for attributed graph embedding to improve the representations' stability and quality. The problem of preserving the structure information is converted into keeping inter-node similarity between non-euclidean high dimensional latent space and euclidean input space. For DMVGAE, firstly, we use a variational auto-encoder mechanism to learn the distribution and get codes from it. Secondly, we propose a graph geodesic similarity to collect graph structure information and node feature information to measure node-to-node relationships in the input and latent space. $t$-distribution is used as a kernel function to fit the neighborhoods between nodes to balance intra-cluster and inter-cluster relationships. Therefore, this method takes advantage of both manifold learning based and auto-encoder-based methods for attributed graph embedding, which is a successful attempt to embed them together, with the reason that we usually think about graphs in terms of their combinatorial properties, variational auto-encoder with the data distribution properties, whereas manifolds in terms of their topological and geometric properties. Thus, our contributions can be summarized as follows:
\begin{itemize}
\item Obtain the topological and geometric properties of graph data under a pre-defined distribution, improve the learned representations' stability and quality, and tackle the crowding problem.
\item Propose a manifold learning loss that considers graph structure information and node feature information to preserve the observing node-to-node geodesic similarity.
\item Achieve state-of-the-art performance on different tasks on benchmark through extensive experiments.
\end{itemize}

\section{Method}
\label{sec:format}

\subsection{Problem Statement}
An attributed graph is denoted as $G=(V,E,X)$, where $V=\{v_{1}, v_{2},\cdots ,v_{n}\}$ is the vertex set with $n$ nodes, $E$ is the edge set, and $X=[x_{1},x_{2},\cdots,x_{n}]^{T}$ is the feature matrix. Compared with original graph $G$, attributed graph $\bar{G}=(V,\bar{E}, X)$ is a fully connected graph with no unconnected nodes and $\bar{E}$ is the new edge set. The topology structure of graph $G$ can be denoted by adjacency matrix $A$:
\begin{equation}
    A=\{a_{ij}\}\in \mathbb{R} ^{n\times n}, 
    a_{ij}=1 \ \mathrm{if} \ (v_{i},v_{j})\in E \ \mathrm{else} \ 0
\label{(1)}
\end{equation} 
$a_{i,j}=1$ indicates that there is an edge from node $v_{i}$ to node $v_{j}$. We aim to find a set of low-dimensional embeddings $Z=[z_{1},z_{2},\cdots,z_{n}]^{T}$ which can relieve and even tackle the crowding problem via preserving both the local and global topological features, the resulting embeddings $Z$ can be used to accomplish a variety of downstream tasks like node clustering, link prediction, and visualization.

\subsection{Proposed method}
\label{method}
To increase the nonlinearity of node features, we use $L$ Fully-Connected (FC) layers to transform the input feature $X$ into $X'$. And then a two-layer GCN is adopted as the graph encoder like VGAE, which takes a Gaussian prior $p(Z)={\textstyle \prod_{i}} p(z_{i}) ={\textstyle \prod_{i}} \mathcal{N}(z_{i}|0,I) $. Generating an approximation $q(Z|X',A)$ of the posterior probability by the variational graph encoder,  we optimize the variational lower bound:
\begin{equation}
L_{0}=\mathbb{E}_{q(Z|(X',A))} [\mathrm{log}\ p(\hat{A}|Z)]-KL[q(Z|X',A)||p(Z)]
\label{(2)}
\end{equation}
where $KL[q(\cdot )||p(\cdot)]$ is the Kullback-Leibler divergence between $q(\cdot)$ and $p(\cdot)$.

For the non-probabilistic graph auto-encoder, we minimize the reconstruction error of the graph data by:
\begin{equation}
L_{1}=\mathbb{E}_{q(Z|(X',A))} [\mathrm{log}\ p(\hat{A}|Z)]
\label{(3)}
\end{equation}
where $\hat{A}$ is the reconstructed graph.

In order to relieve and even tackle the crowding problem, we introduce DML to preserve the geometric structure of the graph $G$. In detail, we calculate graph geodesic distance matrices on prior graph $G_X$ and on complete graph $\bar{G}_X$ in the input space before training, where the prior graph is the given graph or $k$-nearest graph $G_X=G=(V, E, X)$, and the complete graph $\bar{G}_{X}=(V,\bar{E}, X)$, as we want to get local $(G_X)$ and global $(\bar{G}_{X})$ structure features from different aspects. In experiments, we find that if we only use the given graph or $k$-nearest graph $G_X$, the clustered nodes of the same class still easily get together. In the latent space, we sample $K$ latent embeddings $(Z_0, Z_1, \cdots, Z_i, \cdots, Z_K)$ from $q(Z|X',A)$ and calculate graph geodesic distance matrices only on local graphs $G_{Z_i}=(V,E, Z_i)$ in order to reduce the algorithm complexity and training time. The details are shown in Algorithm.\ref{alg:algorithm}. These operations differ from other DML-based embedding methods, which contribute to tackling the crowding problem.  

for a given graph $G_X$, we define the graph geodesic distance matrix  $D(G_X)=\left \{ d^{G_X}_{ij}|i,j=1,2, \cdots,n  \right \}$, where $d^{G_X}_{ij}$ is calculated from Euclidean distances of $(x_i, x_j)$ when $(v_i,v_j)\in E$, otherwise, $d^{G_X}_{ij}$ are set to be a large number. Similarly, we can get $D(\bar{G}_X)$ and $D(G_{Z_i})$.

In order to avoid the adverse effects of outliers and neighborhood inhomogeneity on the characterization of the manifold, the distance $d_{ij}$ is preprocessed to $d_{i|j}$:
\begin{equation}
d_{i|j}={d_{ij}-\rho _{i}}
\label{(4)}
\end{equation}
where $\rho _{i}=\mathrm{min}([d_{i0},d_{i1},\cdots ,d_{in}]) $ is deducted from distances of all others nodes to node $v_{i}$ for alleviating the influence of outliers. After getting a distance matrix, we can use a non-linear function convert the preprocessed distance matrix to a similarity matrix. Different from t-SNE~\cite{LaurensvanderMaaten2008VisualizingDU} and UMAP~\cite{mcinnes2018umap}, the former uses the normalized Gaussian and Cauchy functions and the latter uses the fitted polynomial function, here we adopt $t$-distribution as we find that the degree of freedom $\nu$ in $t$-distribution can be used as a tool to prevent the training from converging to bad local minima and control the separation margin between different manifolds in experiments which means that changing $\nu$ can relieve the crowding problem. The similarity is formulated as follows:
\begin{equation}
\begin{aligned}
p_{i|j}(\sigma _{i},\nu) &= g(d_{i|j},\sigma _{i},\nu) \\
 &=C_{\nu}(1+\frac{d_{i|j}}{\sigma _{i}\nu} )^{-\frac{(\nu+1)}{2} } 
\label{(5)}
\end{aligned}
\end{equation}
\begin{equation}
C_{\nu}= \sqrt{2\pi} \frac{\Gamma (\frac{\nu+1}{2} )}{\sqrt{\nu\pi}\Gamma(\frac{\nu}{2} )}
\label{(6)}
\end{equation}
where the degree of freedom $\nu \in \mathbb{R}_{+}$, $\Gamma(\cdot)$ in the coefficient $C_{\nu}$ is the gamma function, and the data-adaptive parameter $\sigma_i>0$, is estimated by a binary search method as in the UMAP:
\begin{equation}
\sum_{j \neq i} p_{i|j}(\sigma _{i},\nu) = \log _2 Q_p
\label{(7)}
\end{equation}
where $Q_p$ is a hyperparameter that controls the compactness of neighbors.

We can get the graph geodesic similarity:
\begin{equation}
P = \{p_{ij}|i,j=1,2,\cdots, n\}
\label{(8)}
\end{equation}
where $p_{ij}$ is the joint probability by symmetrizing the graph geodesic similarity $p_{i|j}$, which is performed by: 
\begin{equation}
p_{ij} = p_{i|j} + p_{j|i}-2p_{i|j}p_{j|i}
\label{(9)}
\end{equation}

Using above equations, we can calculate graph geodesic distance and get node similarity matrices $P({G_{X}}),P({\bar{G}_{X}})$ and $P({G_{Z_i}})$ on different graphs $G_X=G=(V, E, X)$, $\bar{G}_{X}=(V,\bar{E}, X)$ and $G_{Z_i}=(V,E, Z_i)$ to preserve local and global structure features from different perspectives by a manifold learning loss which is different from graph embedding methods that only take features or structures as input without information maintaining. Here, we adopt a logistic loss:
\begin{equation}
L_{\mathcal{M}}(a,b)=a\log{\frac{a}{b}+(1-a)\log{\frac{1-a}{1-b} } } 
\label{(10)}
\end{equation}
therefore, the loss for structure-preserving is constructed as follows:
\begin{equation}
\begin{aligned}
L_2 =& \frac{1}{K} \sum_{i=0}^{K} (L_{\mathcal{M} }(P({G_{X}}),P({G_{Z_i}})) + \\
&\alpha L_{\mathcal{M} }(P({\bar{G}_{X}}),P({G_{Z_i}}))) 
\label{(11)}
\end{aligned}
\end{equation}
where $\alpha$ is a  weight hyper-parameter, $K$ is the number of samples.

Manifold learning loss $L_{2}$ has ability to tackle the crowding problem as the distance of each pair of nodes is equally calculated on the fully-connected graph $\bar{G}_X$, while distances of the given graph $G_X$ is only performed on the neighbouring nodes, which are both expected to be preserved with distances on graph $G_{Z_i}$ in the latent space. Because $L_{2}$ tries intra-manifold points to transform into a cluster in the latent space and pushes away inter-manifold point pairs from each other. The final loss $L$ (DMVGAE) and $L'$ (DMGAE) is combined as:
\begin{equation}
L=L_2+\beta {L_0}
\label{(12)}
\end{equation}
and 
\begin{equation}
L'=L_2+\beta {L_1}
\label{(13)}
\end{equation}
where $\beta$ is used to balance the reconstruction loss and the manifold learning loss.

We use the prior graph instead of the fully-connected graph in the latent space as well as a well-established mini-batch-based training method to reduce training complexity. Thus, the complexity is $O(KB_{s}N_{n})$ with $N_{n}$ neighbours ($N_{n} < n$), where $B_s$ is the batch size.
\begin{algorithm}[tb]
\caption{DMVGAE/DMGAE}
\label{alg:algorithm}
\begin{algorithmic}
    \State {\bfseries Input:} Graph with links and features:$G[V,E,X]$, weight     hyper-parameters: $\alpha$, $\beta$, number of FC layers: $L$,      number of samples : $K$, learning rate: $l_{r}$, batch size:       $B_{s}$, epochs: $T$, perplexity: $Q_p$, degree of freedom:        $\nu$
    \State {\bfseries Output:} Graph Embedding $Z$
    \State Initialization
    \State Calculate $D({G_X})$ and $D({\bar{G}_X})$, transform them to $P({G_X})$ and $P^(\bar{G}_X)$ by Eq.\ref{(4)}-Eq.\ref{(9)}
 \While{$t=0,1,\cdots,T-1$}
 \State Get $X'$ from the input data $X$ by FC layers
 \State Generate an approximation $q(Z|X',A)$ of the posterior
 probability by the variational graph encoder like VGAE
 \State Calculate loss $L_0$ by Eq.\ref{(2)} or loss $L_1$ by Eq.\ref{(3)} and generate $K$ latent embeddings $Z_0,Z_1,\cdots,Z_K$ from $q(Z|X',A)$
 \While{$k=0,1,\cdots,K-1$}
 \State Calculate $D(G_{Z_{i}})$ and transform it to $P({G_{Z_{i}}})$ by Eq.\ref{(4)}-Eq.\ref{(9)}
 \State Get $L_{\mathcal{M} }(P({G_{X}}),P({G_{Z_i}})), L_{\mathcal{M} }(P({\bar{G}_{X}}),P({G_{Z_i}}))$ by Eq.\ref{(10)}
 \EndWhile
 \State Calculate Loss $L_2$ by Eq.\ref{(11)} 
 \State Calculate Loss $L$ by Eq.\ref{(12)} or Loss $L'$ by Eq.\ref{(13)}
 \State Update network parameters with its stochastic gradient
 \EndWhile
\end{algorithmic}
\end{algorithm}

\section{Experiments}
\subsection{Datasets and settings} Our experiments are conducted on four popular benchmark datasets: Cora, CiteSeer, PubMed, and Wiki. For all datasets, the degree of freedom  $\nu$ in the input space is set to 100. We directly set $\sigma_{i}=1$ and $\rho_{i}=0$ in the latent space as a trade-off between the speed and performance, the necessity of calculating them additionally becomes not so much in the latent space as the layer goes deeper after some nonlinear manifold unfolding. All codes are implemented using the PyTorch library and run on NVIDIA v100 GPU. The best results for each indicator are shown in bold.\\
\textbf{Baselines and metrics.} 
We compare our model with several prevalent and concurrent algorithms: DeepWalk~\cite{perozzi2014deepwalk}, K-means~\cite{WanLeiZhao2018kmeansAR}, various DML and graph embedding methods, GAE and VGAE \cite{kipf2016variational}, DGI \cite{velickovic2019deep}, ARGA \cite{pan2019learning}, AGE \cite{cui2020adaptive}, GIC \cite{Mavromatis2021GraphIM}, MGAE \cite{ChunWang2017MGAEMG}, PANE~\cite{RenchiYang2020ScalingAN}. We employ three metrics for node clustering: Accuracy (ACC), Normalized Mutual Information (NMI), and balanced F1-score (F1). For link prediction, we partition datasets following AGE \cite{cui2020adaptive}, and report Area Under Curve (AUC), and Average Precision (AP).
\subsection{Results on Node clustering}
\begin{table*}[tp]
	\centering
	\caption{Experimental results of node clustering.}
    \setlength{\tabcolsep}{1.5mm}{\begin{tabular}{cc|ccc|ccc|ccc|ccc}
    \toprule
    \multirow{2}{*}{Methods} & \multirow{2}{*}{Input}&\multicolumn{3}{c|}{Cora} & \multicolumn{3}{c|}{Citeseer} & \multicolumn{3}{c|}{PubMed}& \multicolumn{3}{c}{Wiki} \\
    &&ACC&NMI&F1&ACC&NMI&F1&ACC&NMI&F1&ACC&NMI&F1\\
    \toprule
    K-means&Feature&0.347&0.167&0.254&0.385&0.170&0.305&0.573&0.291&0.574&0.334&0.302&0.245\\
    DeepWalk&Graph&0.467&0.318&0.381&0.362&0.970&0.267&0.619&0.167&0.471&0.385&0.324&0.257\\
    \midrule
    GAE&Both&0.533&0.407&0.420&0.413&0.183&0.291&0.641&0.230&0.493&0.173&0.119&0.154\\
    VGAE&Both&0.560&0.385&0.415&0.444&0.227&0.319&0.655&0.251&0.510&0.287&0.303&0.205\\
    MGAE&Both&0.634&0.456&0.380&0.636&0.398&0.395&0.439&0.082&0.420&0.501&0.480&0.392\\
    DGI&Both&0.713&0.564&0.682&0.688&0.444&0.657&0.533&0.181&0.186&-&-&-\\
    ARGA&Both&0.640&0.449&0.619&0.573&0.350&0.546&0.591&0.232&0.584&0.414&0.395&0.383\\
    AGE&Both&0.712&0.559&0.682&0.569&0.348&0.544&-&-&-&0.519&0.494&0.408\\
    GIC&Both&0.725&0.537&0.694&0.696&0.453&0.654&0.673&0.319&0.704&0.505&0.486&0.438\\
    \midrule
    Ours(DMGAE)&Both&0.741&0.578&\textbf{0.703}&0.698&0.452&0.666&0.752&\textbf{0.384}&0.760&0.534&0.493&\textbf{0.479}\\
    Ours(DMVGAE)&Both&\textbf{0.745}&\textbf{0.584}&0.702&\textbf{0.701}&\textbf{0.459}&\textbf{0.668}&\textbf{0.758}&0.382&\textbf{0.765}&\textbf{0.538}&\textbf{0.511}&0.476\\
    \bottomrule 
    \end{tabular}
    }
    \label{table2}
\end{table*}
In the node clustering task, the generated embeddings are clustered into several clusters by the K-means algorithm in an unsupervised manner, which are then evaluated by true external labels. Results are shown in Table \ref{table2}. In order to get a fair comparison, we used K-means for the embeddings of AGE. Algorithms, whether they use feature and graph information or not, are presented. We can see that DMVGAE outperforms almost all of these state-of-the-art methods on the four datasets. The results of DMVGAE are much better than those of VGAE, which indicates that the manifold learning loss $L_2$ is essential to get better graph embeddings for node clustering.
\subsection{Link Prediction Results}
In the link prediction task, some edges are hidden randomly in the input
graph and the goal is to predict the existence of hidden edges based on the computed embeddings. We follow the setup in \cite{kipf2016variational}. Results are shown in Table \ref{table3}. Our proposed method achieves the highest average values on these datasets. 
\begin{table}[htbp]
	\centering
	\caption{Link prediction performance on Cora, Citeseer, and PubMed.}
    \setlength{\tabcolsep}{0.5mm}{\begin{tabular}{c|cc|cc|cc}
    \toprule
    \multirow{2}{*}{Methods}&\multicolumn{2}{c|}{Cora}&\multicolumn{2}{c|}{Citeseer}&\multicolumn{2}{c}{PubMed}\\
    &AUC&AP&AUC&AP&AUC&AP\\
    \midrule  
    DeepWalk&0.831&0.850&0.805&0.836&0.844&0.841\\
    VGAE&0.914&0.926&0.980&0.920&0.964&0.965\\
    GIC&0.935&0.933&0.970&0.968&0.937&0.935\\
    AGE&0.924&0.932&0.924&0.930&0.968&0.971\\
    PANE & 0.933 & 0.918 & 0.932 & 0.919 & \textbf{0.985} & \textbf{0.977} \\
    Ours(DMGAE)&0.966&0.961&0.979&\textbf{0.981}&0.965&0.947\\
    Ours(DMVGAE)&\textbf{0.968}&\textbf{0.977}&\textbf{0.981}&0.978&0.968&0.966\\
    \bottomrule 
    \end{tabular}
    }
    \label{table3}
\end{table}
\subsection{Visualization}
\begin{figure}[htbp!]
\centering
\subfigure[GIC]{\includegraphics[width=2.5cm]{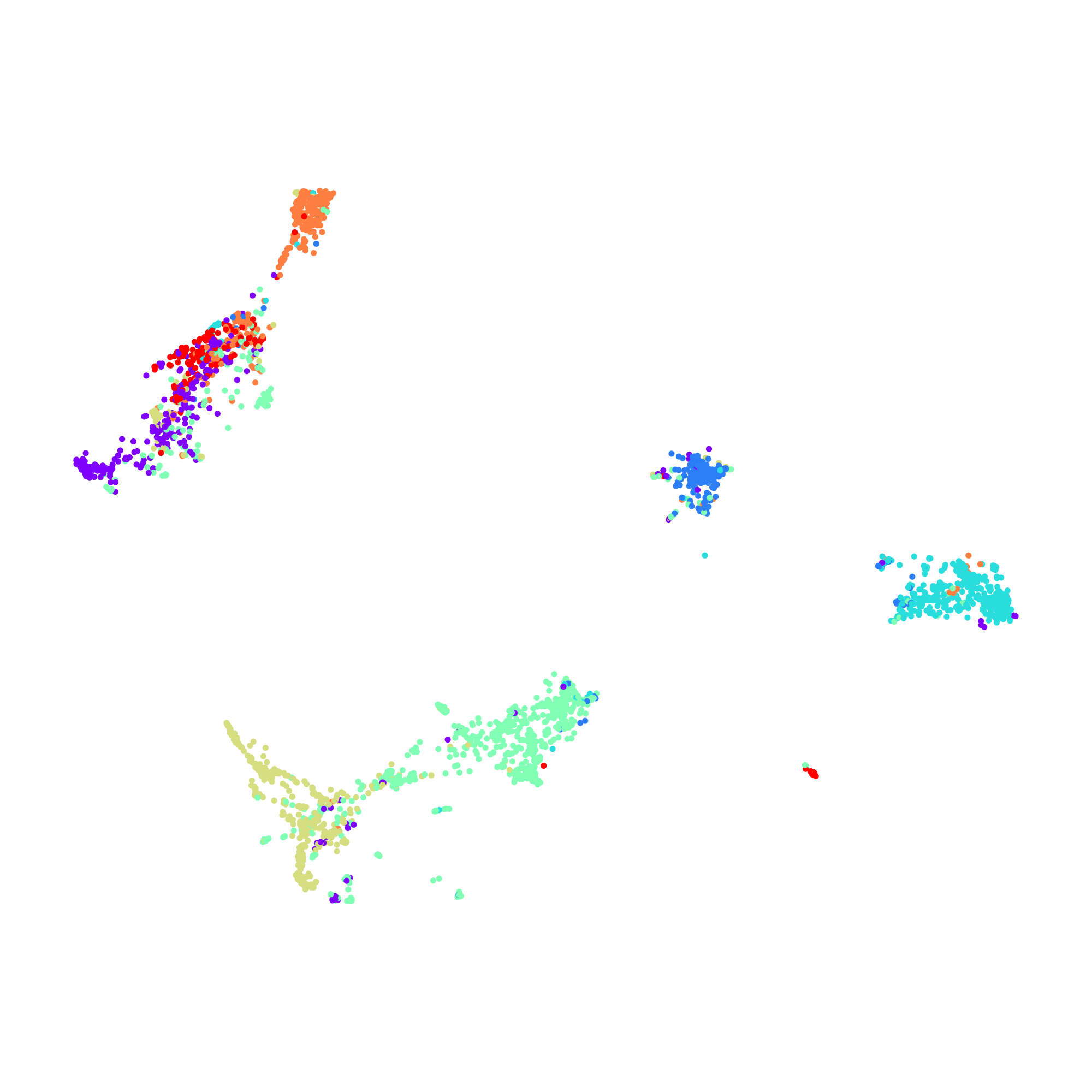}}
\subfigure[AGE]{\includegraphics[width=2.5cm]{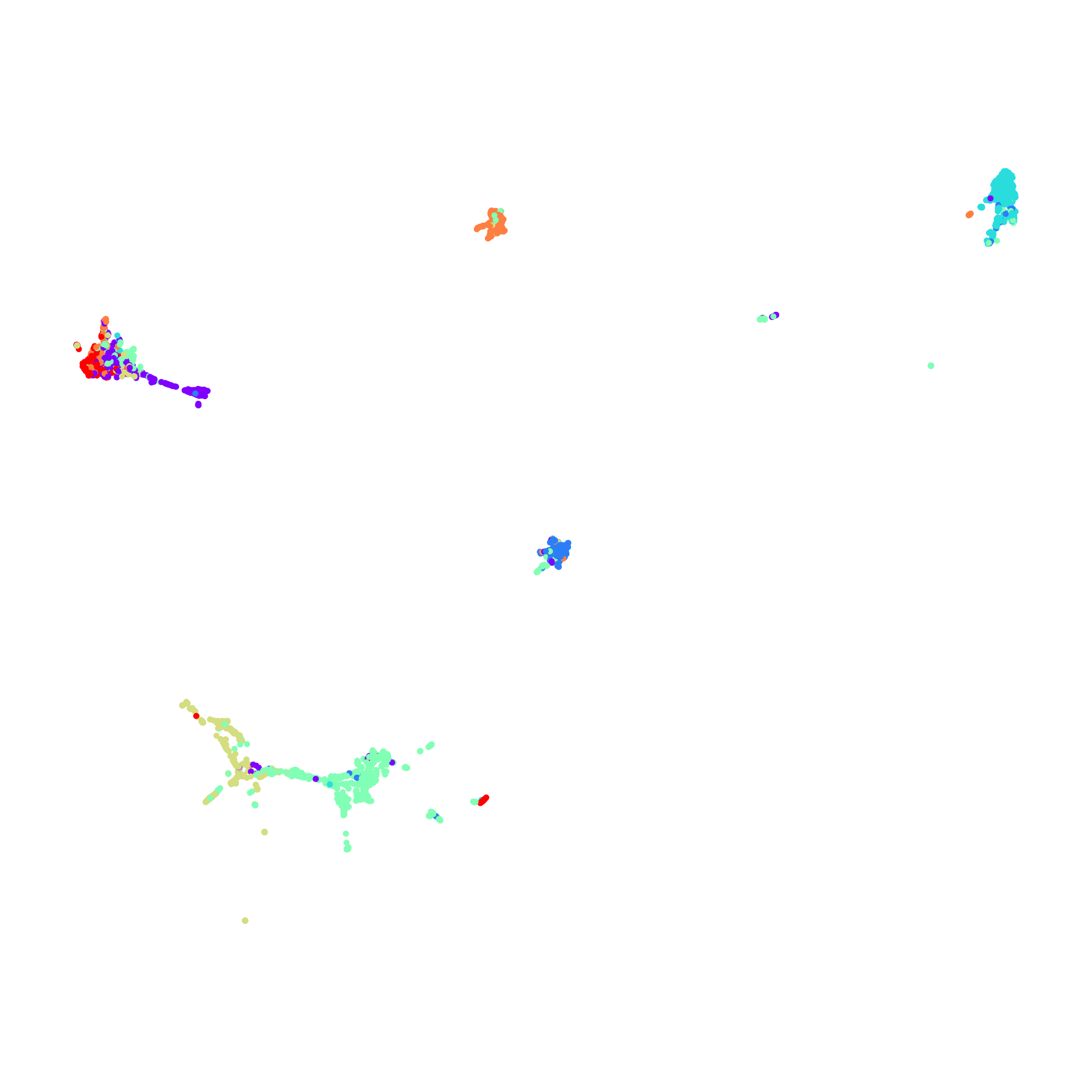}}
\subfigure[Proposed]{\includegraphics[width=2.5cm]{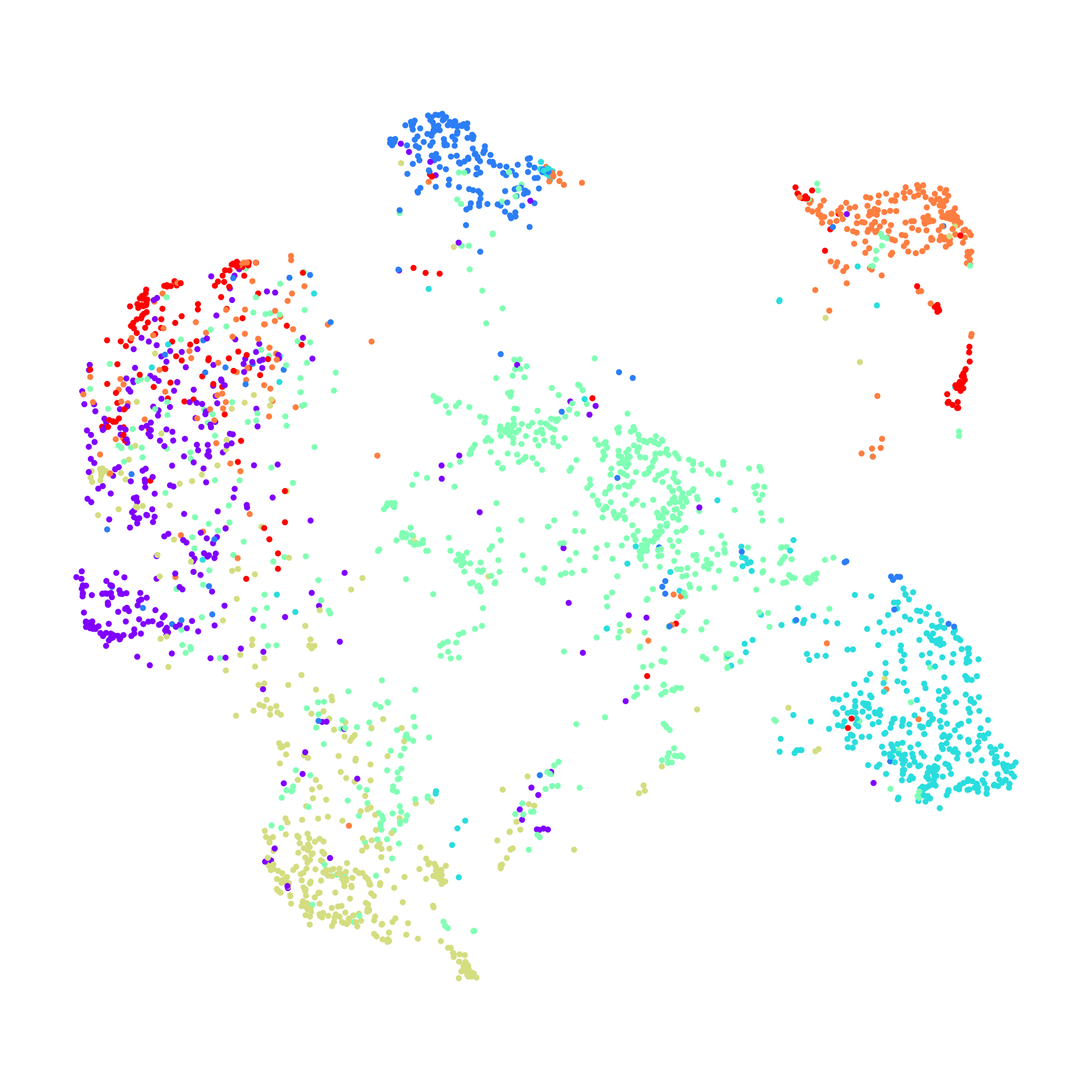}}
\caption{Comparison of visualization results on Cora using UMAP. Colors represent label categories.}
\label{fig3}
\end{figure}

To evaluate our proposed method, we visualize the distribution of the learned latent representations on Cora compared to each node's input features in two-dimensional space using UMAP. As shown in Fig. \ref{fig3}, GIC and AGE suffer from a more severe crowding problem, and our proposed method tackles this problem, performing much better. 
\section{Conclusion}
This paper proposed a deep manifold (variational) graph auto-encoder method (DMVGAE/DMGAE) for attributed graph embedding. 
We introduce DML to graphs and try to preserve the structure information to tackle the crowding problem for the learned embeddings. 
Experiments on standard benchmarks demonstrate our solution. For future work, we plan to introduce different types of noise in the given graph, which is significant in real life to prevent attacks and improve model robustness.

\section{ACKNOWLEDGMENTS}
This work is supported by the Science and Technology Innovation 2030- Major Project (No. 2021ZD0150100) and National Natural Science Foundation of China (No. U21A20427) and Westlake University Funded
Scientific Research Project (No. WU2022C043).

\vfill\pagebreak
\bibliographystyle{IEEEbib}
\bibliography{strings,refs}

\begin{thebibliography}{10}

\bibitem{hastings2006community}
Matthew~B Hastings,
\newblock ``Community detection as an inference problem,''
\newblock {\em Physical Review E}, vol. 74, no. 3, pp. 035102, 2006.

\bibitem{kipf2016semi}
Thomas~N Kipf and Max Welling,
\newblock ``Semi-supervised classification with graph convolutional networks,''
\newblock {\em arXiv preprint arXiv:1609.02907}, 2016.

\bibitem{neyshabur2013netal}
Behnam Neyshabur, Ahmadreza Khadem, Somaye Hashemifar, and Seyed~Shahriar Arab,
\newblock ``Netal: a new graph-based method for global alignment of
  protein--protein interaction networks,''
\newblock {\em Bioinformatics}, vol. 29, no. 13, pp. 1654--1662, 2013.

\bibitem{wang2017mgae}
Chun Wang, Shirui Pan, Guodong Long, Xingquan Zhu, and Jing Jiang,
\newblock ``Mgae: Marginalized graph autoencoder for graph clustering,''
\newblock in {\em CIKM}, 2017, pp. 889--898.

\bibitem{DCRN}
Yue Liu, Wenxuan Tu, Sihang Zhou, Xinwang Liu, Linxuan Song, Xihong Yang, and
  En~Zhu,
\newblock ``Deep graph clustering via dual correlation reduction,''
\newblock in {\em AAAI}, 2022, vol.~36, pp. 7603--7611.

\bibitem{kipf2016variational}
Thomas~N Kipf and Max Welling,
\newblock ``Variational graph auto-encoders,''
\newblock {\em arXiv preprint arXiv:1611.07308}, 2016.

\bibitem{JunXia2023SimGRACEAS}
Jun Xia, Lirong Wu, Jintao Chen, Bozhen Hu, and Stan Z.Li,
\newblock ``Simgrace: A simple framework for graph contrastive learning without
  data augmentation,''
\newblock 2023.

\bibitem{xia2022progcl}
Jun Xia, Lirong Wu, Ge~Wang, and Stan~Z. Li,
\newblock ``Progcl: Rethinking hard negative mining in graph contrastive
  learning,''
\newblock in {\em International conference on machine learning}. PMLR, 2022.

\bibitem{newman2006finding}
Mark~EJ Newman,
\newblock ``Finding community structure in networks using the eigenvectors of
  matrices,''
\newblock {\em Physical review E}, vol. 74, no. 3, pp. 036104, 2006.

\bibitem{li2018community}
Ye~Li, Chaofeng Sha, Xin Huang, and Yanchun Zhang,
\newblock ``Community detection in attributed graphs: An embedding approach,''
\newblock in {\em AAAI}, 2018.

\bibitem{perozzi2014deepwalk}
Bryan Perozzi, Rami Al-Rfou, and Steven Skiena,
\newblock ``Deepwalk: Online learning of social representations,''
\newblock in {\em Proceedings of the 20th ACM SIGKDD international conference
  on Knowledge discovery and data mining}, 2014, pp. 701--710.

\bibitem{zhang2019attributed}
Xiaotong Zhang, Han Liu, Qimai Li, and Xiao-Ming Wu,
\newblock ``Attributed graph clustering via adaptive graph convolution,''
\newblock {\em arXiv preprint arXiv:1906.01210}, 2019.

\bibitem{wang2019attributed}
Chun Wang, Shirui Pan, Ruiqi Hu, Guodong Long, Jing Jiang, and Chengqi Zhang,
\newblock ``Attributed graph clustering: A deep attentional embedding
  approach,''
\newblock {\em arXiv preprint arXiv:1906.06532}, 2019.

\bibitem{pan2019learning}
Shirui Pan, Ruiqi Hu, Sai-fu Fung, Guodong Long, Jing Jiang, and Chengqi Zhang,
\newblock ``Learning graph embedding with adversarial training methods,''
\newblock {\em IEEE transactions on cybernetics}, vol. 50, no. 6, pp.
  2475--2487, 2019.

\bibitem{zhou2020graph}
Jie Zhou, Ganqu Cui, Shengding Hu, Zhengyan Zhang, Cheng Yang, Zhiyuan Liu,
  Lifeng Wang, Changcheng Li, and Maosong Sun,
\newblock ``Graph neural networks: A review of methods and applications,''
\newblock {\em AI Open}, vol. 1, pp. 57--81, 2020.

\bibitem{ZhenyuHou2022GraphMAESM}
Zhenyu Hou, Xiao Liu, Yukuo Cen, Yuxiao Dong, Hongxia Yang, Chunjie Wang, and
  Jie Tang,
\newblock ``Graphmae: Self-supervised masked graph autoencoders,''
\newblock {\em knowledge discovery and data mining}, 2022.

\bibitem{van2008visualizing}
Laurens Van~der Maaten and Geoffrey Hinton,
\newblock ``Visualizing data using t-sne.,''
\newblock {\em Journal of machine learning research}, vol. 9, no. 11, 2008.

\bibitem{ChunWang2017MGAEMG}
Chun Wang, Shirui Pan, Guodong Long, Xingquan Zhu, and Jing Jiang,
\newblock ``Mgae: Marginalized graph autoencoder for graph clustering,''
\newblock {\em conference on information and knowledge management}, 2017.

\bibitem{ZelinZang2022DLMEDL}
Zelin Zang, Siyuan Li, Di~Wu, Ge~Wang, Lei Shang, Baigui Sun, Hao Li, and
  Stan~Z. Li,
\newblock ``Dlme: Deep local-flatness manifold embedding,''
\newblock 2022.

\bibitem{LaurensvanderMaaten2008VisualizingDU}
Laurens van~der Maaten and Geoffrey~E. Hinton,
\newblock ``Visualizing data using t-sne,''
\newblock {\em Journal of Machine Learning Research}, 2008.

\bibitem{mcinnes2018umap}
Leland McInnes, John Healy, and James Melville,
\newblock ``Umap: Uniform manifold approximation and projection for dimension
  reduction,''
\newblock {\em arXiv preprint arXiv:1802.03426}, 2018.

\bibitem{WanLeiZhao2018kmeansAR}
Wan-Lei Zhao, Cheng-Hao Deng, and Chong-Wah Ngo,
\newblock ``k-means: A revisit,''
\newblock {\em Neurocomputing}, 2018.

\bibitem{velickovic2019deep}
Petar Velickovic, William Fedus, William~L Hamilton, Pietro Li{\`o}, Yoshua
  Bengio, and R~Devon Hjelm,
\newblock ``Deep graph infomax.,''
\newblock {\em ICLR}, vol. 2, no. 3, pp. 4, 2019.

\bibitem{cui2020adaptive}
Ganqu Cui, Jie Zhou, Cheng Yang, and Zhiyuan Liu,
\newblock ``Adaptive graph encoder for attributed graph embedding,''
\newblock in {\em Proceedings of the 26th ACM SIGKDD International Conference
  on Knowledge Discovery \& Data Mining}, 2020, pp. 976--985.

\bibitem{Mavromatis2021GraphIM}
Costas Mavromatis and G.~Karypis,
\newblock ``Graph infoclust: Maximizing coarse-grain mutual information in
  graphs,''
\newblock in {\em PAKDD}, 2021.

\bibitem{RenchiYang2020ScalingAN}
Renchi Yang, Jieming Shi, Xiaokui Xiao, Yin Yang, Juncheng Liu, and Sourav~S.
  Bhowmick,
\newblock ``Scaling attributed network embedding to massive graphs,''
\newblock {\em very large data bases}, 2020.

\end{thebibliography}

\end{document}